\documentclass[conference]{IEEEtran}
\usepackage{cite}
\usepackage{amsmath,amssymb,amsfonts}
\usepackage{algorithmic}
\usepackage{multirow}

\usepackage{graphicx}
\usepackage{textcomp}
\usepackage{xcolor}
\usepackage{subcaption}
\def\BibTeX{{\rm B\kern-.05em{\sc i\kern-.025em b}\kern-.08em
    T\kern-.1667em\lower.7ex\hbox{E}\kern-.125emX}}
\usepackage{hyperref}
\hypersetup{
    colorlinks=true,
    allcolors=blue
}

\definecolor{ckgreen}{rgb}{0,0.56,0}

\usepackage{array}
\newcommand{\PreserveBackslash}[1]{\let\temp=\\#1\let\\=\temp}
\newcolumntype{C}[1]{>{\PreserveBackslash\centering}p{#1}}
\newcolumntype{R}[1]{>{\PreserveBackslash\raggedleft}p{#1}}
\newcolumntype{L}[1]{>{\PreserveBackslash\raggedright}p{#1}}
\usepackage{booktabs}
\graphicspath{ {images/} }
\newcommand{\etal}{\textit{et al}.}

\usepackage{tikz}
\def\checkmark{\tikz\fill[scale=0.4](0,.35) -- (.25,0) -- (1,.7) -- (.25,.15) -- cycle;} 

\usepackage[pscoord]{eso-pic}
\newcommand{\placetextbox}[3]{
\setbox0=\hbox{#3}
\AddToShipoutPictureFG*{ \put(\LenToUnit{#1\paperwidth},\LenToUnit{#2\paperheight}){\vtop{{\null}\makebox[0pt][c]{#3}}}
}
}

\placetextbox{.224}{0.055}
{979-8-3503-4602-2/22/\$31.00 \copyright 2022 IEEE}
\begin{document}
\title{An Efficient Transfer Learning-based Approach for Apple Leaf Disease Classification}
\author{

\IEEEauthorblockN{  
    Md. Hamjajul Ashmafee\IEEEauthorrefmark{1},
    Tasnim Ahmed\IEEEauthorrefmark{1},
    Sabbir Ahmed\IEEEauthorrefmark{1},
    Md. Bakhtiar Hasan\IEEEauthorrefmark{1}, 
    Mst Nura Jahan\IEEEauthorrefmark{2},\\ 
    A.B.M. Ashikur Rahman\IEEEauthorrefmark{3}
                }

\IEEEauthorblockA{\IEEEauthorrefmark{1}Department of Computer Science and Engineering, Islamic University of Technology, Gazipur 1704, Bangladesh}

\IEEEauthorblockA{\IEEEauthorrefmark{2} Department of Entomology, Bangabandhu Sheikh Mujibur Rahman Agricultural University, Gazipur 1706, Bangladesh}

\IEEEauthorblockA{\IEEEauthorrefmark{3} Department of ICS, King Fahd University of Petroleum \& Minerals, Dhahran, Saudi Arabia}

\IEEEauthorblockA{
    \{\IEEEauthorrefmark{1}ashmafee,
        \IEEEauthorrefmark{1}tasnimahmed,
        \IEEEauthorrefmark{1}sabbirahmed,
        \IEEEauthorrefmark{1}bakhtiarhasan
        \}@iut-dhaka.edu,
\IEEEauthorrefmark{2}nura.jahan01@gmail.com,\\
\IEEEauthorrefmark{3}g202204800@kfupm.edu.sa}
}

\makeatletter
\let\old@ps@IEEEtitlepagestyle\ps@IEEEtitlepagestyle
\def\confheader#1{%
    \def\ps@IEEEtitlepagestyle{%
        \old@ps@IEEEtitlepagestyle%
        \def\@oddhead{\strut\hfill#1\hfill\strut}%
        \def\@evenhead{\strut\hfill#1\hfill\strut}%
    }%
    \ps@headings%
}
\makeatother

\confheader{%
        \parbox{20cm}{2023 International Conference on Electrical, Computer and Communication Engineering (ECCE)}
}


\maketitle

\begin{abstract}
Correct identification and categorization of plant diseases are crucial for ensuring the safety of the global food supply and the overall financial success of stakeholders. In this regard, a wide range of solutions has been made available by introducing deep learning-based classification systems for different staple crops. Despite being one of the most important commercial crops in many parts of the globe, research proposing a smart solution for automatically classifying apple leaf diseases remains relatively unexplored. This study presents a technique for identifying apple leaf diseases based on transfer learning. The system extracts features using a pretrained EfficientNetV2S architecture and passes to a classifier block for effective prediction. The class imbalance issues are tackled by utilizing runtime data augmentation. The effect of various hyperparameters, such as input resolution, learning rate, number of epochs, etc., has been investigated carefully. The competence of the proposed pipeline has been evaluated on the apple leaf disease subset from the publicly available `PlantVillage' dataset, where it achieved an accuracy of 99.21\%, outperforming the existing works.

\end{abstract}


\section{Introduction}

Infection diseases yearly reduce the potential harvest by an average of 40\% to even 100\% in developing countries \cite{panno2021review}. Traditional monitoring systems often require expert support and have several limitations. Such methods often rely on manual data collection, which is time-consuming and error-prone. Moreover, the process is labor-intensive, which makes it expensive and adds difficulty in the real-time monitoring of large areas of land. These limitations make it difficult for traditional monitoring systems to provide accurate and comprehensive information about conditions in an agricultural field and can limit the effectiveness of decision-making for farmers and agricultural stakeholders.
The lack of agricultural knowledge of crop growers adds to this problem \cite{agriculture11080707}. 
In this regard, intelligent agricultural systems with the ability of fast and accurate disease predictions can go a long way toward ensuring global food security. Different computer vision techniques can be utilized for the early detection of plant disease using leaf images \cite{dhaka2021aSurvey}.

For the categorization of plant diseases, several methods have been put forth utilizing conventional machine learning-based techniques. These works, however, rely on handcrafted feature selection methods that haven't been able to generalize to bigger datasets\cite{s18082674}. 
In this connection, Deep learning (DL) based algorithms have opened up new possibilities for agriculture because of their remarkable generalization abilities, which eliminate the need for intensive feature extraction \cite{kamilaris2018deep}.
These algorithms can automatically learn features from large-scale real-world leaf image datasets containing a wide variety of color, texture, and shape features with high precision.
Moreover, Convolutional Neural Network (CNN) has emerged as an effective method for any classification assignment thanks to its capability to efficiently identify key characteristics from images without expert intervention and adapt to a variety of leaf disease recognition tasks. Furthermore, modern CNN-based architectures like ResNets, DenseNets, InceptionNets, etc. have provided the ability to comprehend complicated patterns, allowing them even to outperform humans in several detection and categorization tasks \cite{kamal2022huruf, alamgir2022dhakaAI, sakib2021bdslReview, bakhtiar2022traffic, bakhtiar2022heatgait}.

By employing a model that is effective at solving one issue as the basis for another, transfer learning has significantly reduced the need for substantial processing resources. In this process, the models are already trained with large-scale image classification datasets; they have the capability to learn almost all the basic features from a given image \cite{khan2022rethinking}. Fine-tuning these pretrained architectures with the domain-specific datasets significantly improves their ability within just a few epochs.
As a result, transfer learning has been used to categorize leaf diseases using pretrained architectures on publicly available datasets, opening the door for a wide range of solutions in the current literature \cite{mohanty2016using}. 
These deep neural networks based on transfer learning have been found to be extraordinarily useful for leaf disease classification tasks for a number of plants such as rice \cite{sethy2020deep}, corn \cite{mishra2020deep}, tomato \cite{ahmed2022less}, potato \cite{tiwari2020potato}, etc. 
However, despite being one of the most significant commercial crops in many regions of the world, these smart solutions still need to be explored in categorizing apple leaf diseases. 

Apple production and planting area have grown steadily over time due to its great economic and nutritional significance. However, overall production is falling because the crop is vulnerable to many different diseases \cite{liu2017identification}. As a result, there is a high need for reliable techniques for apple leaf disease identification. Earlier methods for classifying apple leaf disease included a variety of image-based, manually crafted feature selection methods that were then put into classifiers utilizing machine learning algorithms \cite{ramedani2014potential, chuanlei2017apple}. These studies were frequently restricted to specific contexts and generally concentrated on a small number of diseases with extensive feature engineering. Hence the recent works focused on utilizing different Deep Learning techniques with high generalization ability.

Zhang \textit{et al}. \cite{chuanlei2017apple} proposed a method utilizing color transformation, background removal, and spot segmentation techniques for apple leaf disease classification. The suggested pipeline extracted 38 attributes relating to the color, texture, and form of each spot image. Next the most essential features were selected using a Genetic Algorithm and Correlation-based feature extraction approaches and passed to an SVM-based classifier. For the identification of four types of apple leaf diseases, Liu \textit{et al}. \cite{liu2017identification} designed a pipeline utilizing a pretrained AlexNet architecture, which achieved an accuracy of 97.62\%. In another work, a CNN-based architecture named VGG-INCEP was suggested by Jiang \textit{et al}. \cite{jiang2019real} to distinguish between four apple leaf diseases and to guarantee real-time disease detection. In a different study, the authors used a pretrained DenseNet121 architecture and assessed the model using three loss functions on a dataset that included samples from six different classes of both healthy and diseased samples \cite{zhong2020research}.

Using the apple leaf subset of the PlantVillage dataset, Baranwal \textit{et al.} developed a modified LeNet-based architecture and achieved a promising accuracy of 98.54\% \cite{baranwal2019deep}. On the same dataset, Albayati et al. proposed a Deep CNN that combines accelerated robust feature extraction with grasshopper optimization \cite{al2020evolutionary}. Li \textit{et al.} proposed a technique based on segmentation and transfer learning utilizing pretrained ResNet18 architecture \cite{li2020apple}. Agarwal \textit{et al.} proposed a modified CNN-based architecture adopting different augmentation techniques and achieved 99\% accuracy \cite{agarwal2019fcnn}. However, some of these works did not utilize the entire dataset that might have missed some critical samples, and the performance of these systems can be further improved by utilizing more recent variants of state-of-the-art Deep CNN-based transfer learning approaches. 

This work proposes a classification method for apple leaf diseases that is based on transfer learning. As a feature extractor, the pipeline employs a pretrained EfficientNetV2S architecture, supplemented by a classifier network. In order to solve the problem of class imbalance, efficient methods of augmenting data at runtime have been introduced. The impact of different hyperparameters, such as image resolution, learning rate, number of epochs, etc., have been thoroughly analyzed. Experimental results reveal that the suggested pipeline outperforms the prior work on apple leaf disease classification tasks.

The remaining sections are organized as follows. In Section 2, details of the different components of our proposed methodology are described. Section 3 provides the experimental data justifying the effectiveness of our suggested strategy. Section 4 contains our concluding remarks, and discusses the shortcomings with recommendations for future research.

\section{Methodology}

\begin{figure*}[tb]
    \centering
    \includegraphics[width=0.95\textwidth]{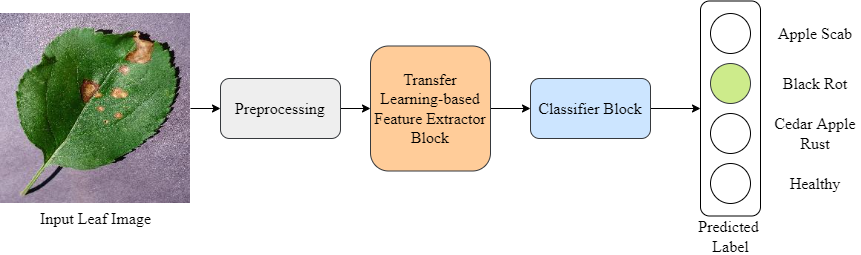}	
    \caption{Overview of the Apple Leaf Disease Classification Pipeline}
    \label{fig:pipeline}
\end{figure*}

The proposed pipeline receives apple leaf images as input and generates the class label as output. First, the images are pre-processed to normalize the pixel values. Then they are passed through a pre-trained feature extractor, EfficientNetV2S \cite{tan2021efficientnetv2} to extract meaningful features. These features are then fed to a shallow densely connected classifier block to determine the class label. The overall pipeline is shown in Figure \ref{fig:pipeline}.

\subsection{Dataset}
\begin{figure}[tb]
    \centering
    \subfloat[Apple Scab]{
    \includegraphics[width=0.2\textwidth, height=3cm]{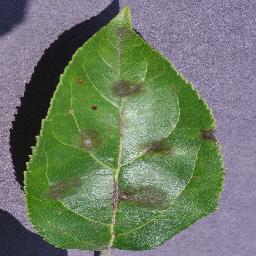}	
    }
    \subfloat[Black Rot]{
    \includegraphics[width=0.2\textwidth, height=3cm]{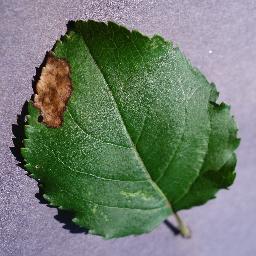}}\\
    
    \subfloat[Cedar Apple Rust]{
    \includegraphics[width=0.2\textwidth, height=3cm]{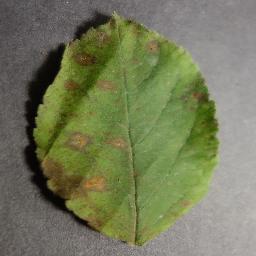}}
    \subfloat[Healthy]{
    \includegraphics[width=0.2\textwidth, height=3cm]{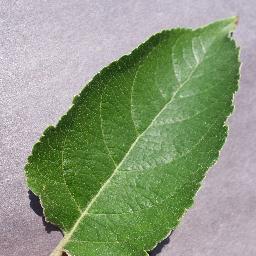} }

    \caption{Apple leaf images from the PlantVillage dataset}
    \label{fig:dataset}
\end{figure}

We have used the apple leaf disease subset of the PlantVillage dataset, one of the largest open-access sources of leaf images \cite{hughes2015open}. There are 54,309 images of healthy and diseased leaves of 14 crops in the plantVillage dataset, all of which have been annotated by experts. There are 3,171 pictures of apple leaves, with 3 groups representing diseases and 1 representing healthy leaves. Figure \ref{fig:dataset}  shows examples of images in each category.

The images were captured in a laboratory setting and depict varying degrees of disease spot distribution on the leaf. Class imbalance is present in the dataset, as the greatest number of 'healthy' samples is 1645, while the number of samples corresponding to 'Cedar apple rust' is only 275. To combat this imbalance issue, we have included the concept of runtime augmentation.

\subsection{Augmentation}
Several augmentation techniques are employed to address the class imbalance issue. Rotation, Horizontal Flipping, Height and Width Shift, and Shearing are performed randomly during the training and testing procedure to emulate the real-life scenario where the pipeline will be implemented \cite{tasnim2019bangla}. These augmentations are randomly combined during training to ensure that the feature extractor and classifier blocks go through different variations of the images in every epoch learning to recognize diverse image samples.

Runtime augmentation is performed, contrasting the conventional augmentation approaches used in leaf disease classification \cite{ahmed2022less}. This is because performing augmentation before splitting the dataset can inject different variations of the same leaf image in training and test set resulting in data leakage \cite{morshed2022fruit}. On the other hand, using runtime augmentation reduces overfitting as the feature extractor and classifier blocks do not go through the same images in each epoch.

\begin{table}[b]
    \centering
    \caption{Configuration of the Feature Extraction Block}
    \label{tab:efficientNetV2S}
    \begin{tabular}{ C{2cm} C{1.2cm} C{0.7cm} C{1.4cm} C{1.2cm}} 
    \toprule
    \textbf{Operator} & \textbf{Filter Size} & \textbf{Stride} & \textbf{Number of Channles} &
    \textbf{Number of Layers}
    \\
    \midrule
    Conv & $3\times3$ & 2 & 24 & 1\\
    Fused-MBConv1 & $3\times3$ & 1 & 24 & 2 \\
    Fused-MBConv4 & $3\times3$ & 2 & 48 & 4 \\
    Fused-MBConv4 & $3\times3$ & 2 & 64 & 4 \\
    MBConv4 & $3\times3$ & 2 & 128 & 6\\
    MBConv6 & $3\times3$ & 1 & 160 & 9 \\
    MBConv6 & $3\times3$ & 2 & 256 & 15\\
    Conv & $1\times1$ & - & 1280 & 1\\
    \midrule
    \end{tabular}
    
\end{table}

\subsection{Feature Extraction}

Conventional machine learning algorithms rely on training and test data from the same feature space  \cite{yasmeen2021csvcNet}. Yet, developments in deep neural networks have made it possible to reuse architectures trained to extract features from training data of one domain to extract features from another domain. This method of propagating knowledge, also known as Transfer Learning, has increased learning performance while reducing the computational requirements for training models from scratch \cite{ashikur2022twoDecades}.

The MBConv operator is taken from MobileNetV2 architecture \cite{Sandler_2018_CVPR}. It consists of three convolutional layers: Expansion, Depthwise Convolution, and Projection. The Expansion Layer applies Pointwise convolution, increasing the number of channels by an expansion factor. The Depthwise Convolution Layer applies $3 \times 3$ convolution per channel working as a filter. The Projection Layer projects the filtered results to higher dimensions generating salient features. It also reduces the number of channels of the input.

The Fused-MBConv operator is similar to MBConv. However, the Depthwise Convolution Layer and the Expansion Layer are replaced with regular $3 \times 3$ convolution. This results in faster and more efficient execution of the models in edge TPUs, often used in embedded systems.

\subsection{Classification}
Rather than using the features extracted by EfficientNetV2S for predicting the class label, a combination of fully-connected operators is used to fine-tune the extracted features. The fully-connected layers consider various combinations of the features extracted by EfficientNetV2 to improve the classification performance. In addition, the fully-connected operators are sandwiched between batch normalization and dropout layers. The use of a batch normalization layer stabilizes the learning process reducing the number of epochs required to train the pipeline \cite{batchnormalization}. The dropout layer helps regularize the learning process reducing overfitting \cite{dropout}.

\section{Results and Discussions}
In this section, we assess the competency of our proposed architecture with respect to the dataset mentioned above for the task of leaf disease classification in Apple.

\subsection{Experimental Setup}
The proposed pipeline was trained on NVIDIA Tesla T4 GPUs with 15GB of virtual memory in the Google Colaboratory environment. The original input images have been scaled to a size of $256 \times 256 \times 3$. The data was divided into three sets, with a $ 60:20:20$ split across the training, validation, and testing data. Due to limitations in the GPU's memory, the training of the model was carried out using a 32-megabyte batch size. Within five epochs, the training session would be discontinued if no substantial improvement was shown. For all tasks, the Adam optimizer was employed with a Learning rate of 0.0001 and a decay value of 0.1 after five patient epochs.


\subsection{Evaluation Metrics}
In our experiments, we have chosen accuracy, precision, recall, and F-1 score as the evaluation criteria for our proposed architecture. Accuracy is a widely used performance metric. However, accuracy is a bit misleading when referring to an imbalanced dataset. An experiment using an imbalanced dataset might achieve very high accuracy despite having low accuracy for the smaller classes. To avoid this issue, we have considered precision and recall. Because precision and recall are about exactness and completeness, respectively, they are biased as well. For this reason, we have considered the F-1 score as another performance metric to evaluate our model. Finally, a better class-wise performance representation can be obtained from the confusion matrix.

\subsection{Performance Analysis}

\begin{table}[tb]
    \centering
    \caption{Performance Analysis for the Proposed Model}
    \begin{tabular}{C{1.5cm} C{0.9cm} C{0.9cm} C{0.9cm} C{0.9cm} C{0.9cm}}
    \toprule
    \textbf{Input Size} & \textbf{Augment} & \textbf{Accuracy} & \textbf{Precision} & \textbf{Recall} & \textbf{F-1 Score}\\    
    \midrule
    $224\times224$ & $\times$     & 95.74 & 95.02 & 94.13 & 94.57\\
    $224\times224$ & \checkmark & 98.78 & 97.23 & 96.57 & 96.89\\
    $256\times256$ & $\times$     & 96.06 & 96.07 & 95.91 & 95.98\\
    $256\times256$ & \checkmark & 99.21 & 99.15 & 99.10 & 99.12\\
    \midrule
    \end{tabular}
    \label{tab:hyperparamTuning}
\end{table}

Our proposed architecture extracts useful feature information from apple leaves using EfficientNetV2S. Extracted features are then classified into four classes using a dense classifier. The proposed pipeline achieves a summit accuracy of 99.21\% on the dataset. To increase the robustness of our model, we have augmented the dataset at every epoch of training. Since the samples in our dataset are generated in a lab environment, they might not replicate real-world scenarios. In reality, sample images might be in a different orientation, distorted, and so on. So, augmentation improves the overall accuracy and robustness of the proposed model. We also observed that training the model with different augmented versions of the same sample images in every epoch increased the performance metric values as well. We have also experimented with different resolutions of sample images. From our experiments, it was evident that higher resolution samples increase the model performance. Detailed results of our experiments with different hyper-parameters are depicted in Table \ref{tab:hyperparamTuning}.

The dataset we used incorporates an issue of class imbalance, which is a challenge in deep learning-based image classification tasks. Since our feature extractor is pretrained on a substantially large Imagenet dataset, the efficient feature extraction helps our proposed architecture handle class imbalance. Table \ref{tab:classwisePerformance} illustrates the class-wise performance and number of test samples per class.

\begin{table}[tb]
    \centering
    \caption{Class-wise Performance on Apple Leaf Disease Classification}
    \label{tab:classwisePerformance}
    \begin{tabular}{lcccc}
    \toprule
    \textbf{Class Name} & \textbf{Precision} & \textbf{Recall} & \textbf{F-1 Score} & \textbf{Support}\\
    \midrule
    Healthy     & 99 & 98 & 99 & 127\\
    Apple Scab  & 99 & 98 & 99 & 126\\
    Black Rot   & 100& 98 & 99 & 56\\
    Cedar Apple Rust & 99 & 100 & 100 & 326\\
    \midrule
         
    \end{tabular}
\end{table}

\begin{figure}[tb]
    \centering
    \includegraphics[width=\columnwidth]{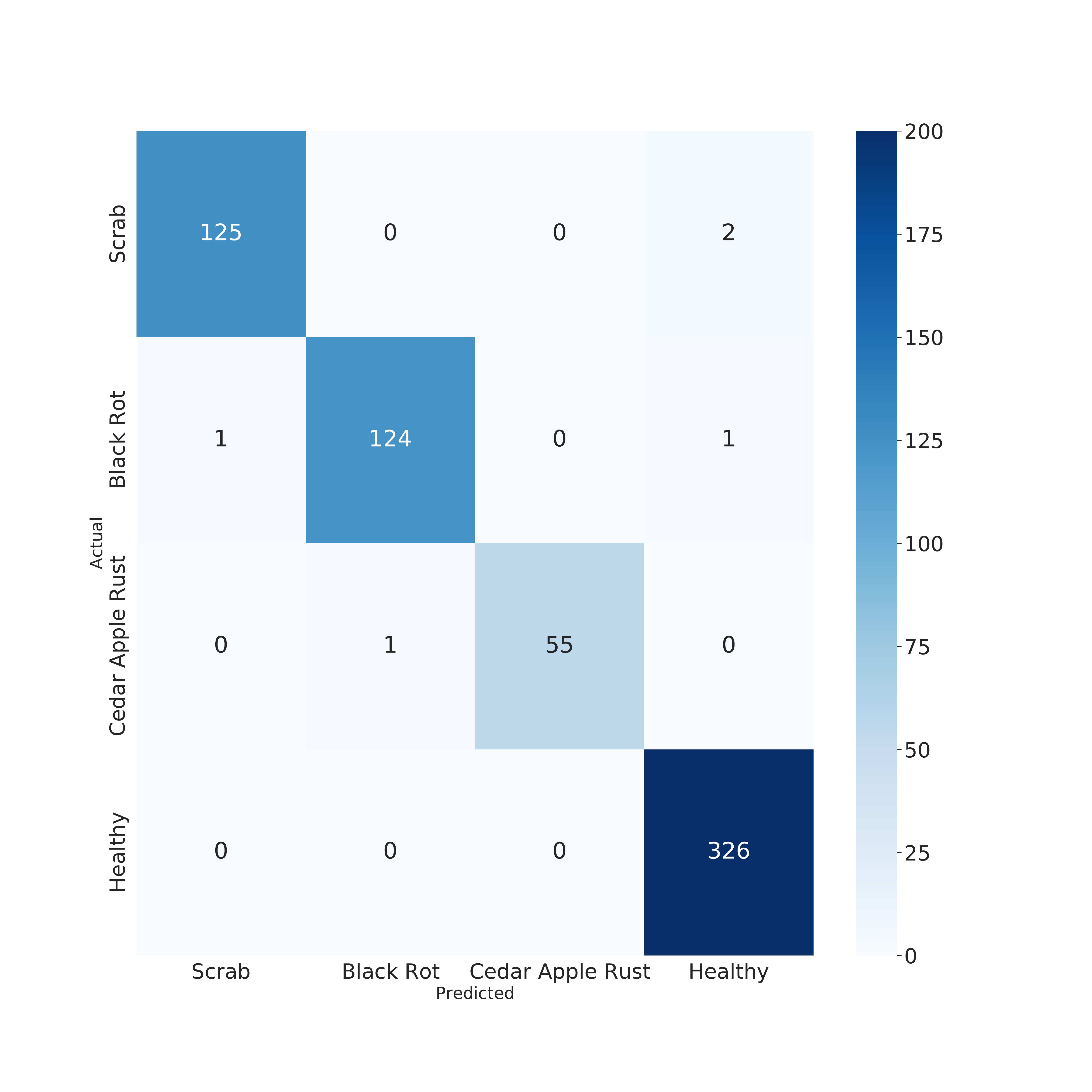}	
    \caption{Confusion Matrix}
    \label{fig:confMat}
\end{figure}

\begin{figure}[htb]
    \centering
    \includegraphics[width=0.4\textwidth]{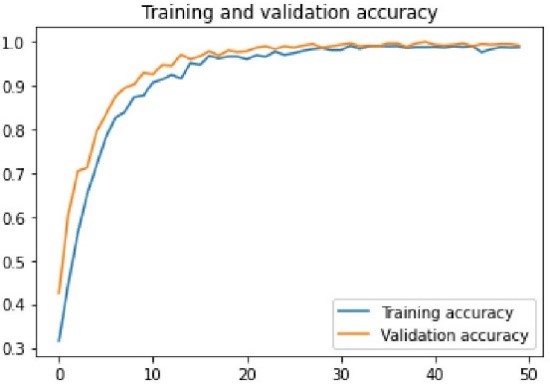}	
    \caption{Training vs Validation Accuracy}
    \label{fig:trainValAccuracy}
\end{figure}

\begin{figure}[tb]
    \centering
    \includegraphics[width=0.4\textwidth]{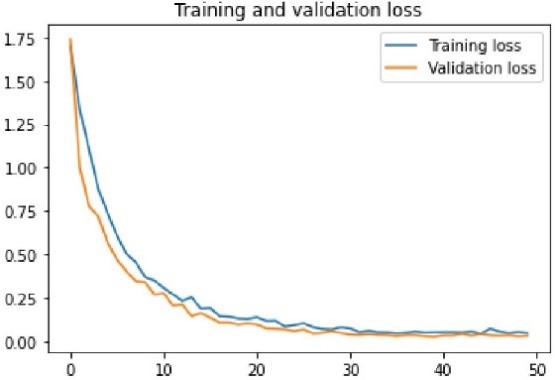}	
    \caption{Training vs Validation Loss}
    \label{fig:trainValLoss}
\end{figure}

Figure \ref{fig:confMat} illustrates the confusion matrix obtained from our best-performing model. It is evident from the performance of our proposed architecture that it performed very well despite having a high-class imbalance in the dataset, thus generalizing pretty well. A dataset with a lower number of samples per class and an imbalanced class distribution often leads to over-fitting of the model. The training and validation accuracy and loss curves shown in \figureautorefname~\ref{fig:trainValAccuracy} and \ref{fig:trainValLoss} illustrate that our proposed architecture successfully overcomes the over-fitting issue by incorporating optimized hyper-parameters as well as dropout layers in the classifier network.

\subsection{Comparison with Existing Works}

Table \ref{tab:sotaCompare} comprises the performance of the existing works related to apple leaf disease classification on the PlantVillage dataset. It is observed that our proposed architecture performs better than the existing works in terms of accuracy. The other metrics we have shown to evaluate our model in the previous subsection are also satisfactory. This performance also proves the efficiency of our feature extractor, EfficientNetV2S, for apple leaf disease classification.

\begin{table}[h]
    \centering
    \caption{Performance Comparison with Existing Works}
    \begin{tabular}{l c}
    \toprule
    \textbf{Reference} & \textbf{Accuracy (\%)}\\
    \midrule
    Al-bayati \etal \cite{al2020evolutionary}    & 98.28\\
    Li \etal \cite{li2020apple}          & 98.50\\
    Baranwal \etal \cite{baranwal2019deep}     & 98.54 \\ 
    Agarwal \etal \cite{agarwal2019fcnn}      & 99.00 \\
    \midrule
    Ours                    & 99.21\\
    \bottomrule
    \end{tabular}
    \label{tab:sotaCompare}
\end{table}


\subsection{Error Analysis}
\begin{figure}[tb]
    \centering
    \subfloat[Cedar Apple Rust affected leaf misclassified as Black Rot\label{subfig:cedar}]{
	\includegraphics[width=.47\columnwidth]{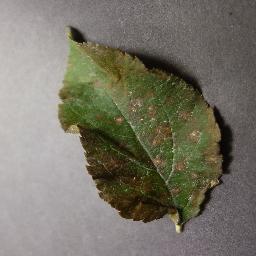}
 }
 \hfill
 \subfloat[Similar Leaf with Black Rot\label{subfig:black}]{
	\includegraphics[width=.47\columnwidth]{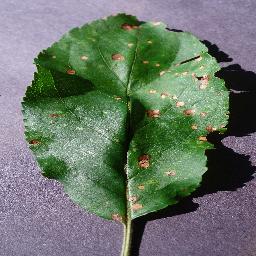}
 }
 \caption{Misclassified sample with a visually similar sample of the predicted class}
 \label{fig:mis}
\end{figure}

As illustrated in \figureautorefname~\ref{fig:confMat}, our proposed pipeline was able to classify almost all the samples correctly. However, there are still some misclassified samples. This can be attributed to the high inter-class similarity between the disease classes. For example, as shown in \figureautorefname~\ref{fig:mis}(\subref{subfig:cedar}), a leaf affected with Cedar Apple Rust was predicted to be affected by Black Rot. Nonetheless, leaves affected with Black Rot may look similar to that of Cedar Apple Rust (\figureautorefname~\ref{fig:mis}(\subref{subfig:black})). The problem is further deteriorated by the fact that the number of training samples for Black Rot is lesser than the other classes. As a result, the model used in our pipeline does not get a good look at Black Rot leaves to extract meaningful features and differentiate them from other disease classes. More image samples of leaves affected by Black Rot could be used to train the model to resolve the issue.

\section{Conclusion}

Early detection and classification are necessary to meet the food demand of the fast-growing population. In summary, this work presents a novel transfer learning-based approach for accurately identifying and categorizing apple leaf diseases. The proposed solution leverages the power of deep transfer learning techniques to extract informative features and make predictions with high accuracy. The system has been thoroughly evaluated on the apple leaf disease subset from the PlantVillage dataset, where it outperformed existing works with a promising accuracy of 99.21\%. The study highlights the effectiveness of transfer learning and data augmentation in handling class imbalance issues and the importance of careful hyperparameter selection for optimal performance. 
Additionally, incorporating domain-specific knowledge into the system can further improve its accuracy and robustness. 
In the future, this approach can be adapted for real-life scenarios meeting the demands of the end users, and the proposed pipeline can be extended to recognize other plant diseases. This work can contribute to ensuring the safety of the global food supply and the financial success of stakeholders by providing a reliable and accurate tool for classifying apple leaf diseases.


\bibliographystyle{IEEEtran}
\bibliography{main}

\begin{thebibliography}{10}
\providecommand{\url}[1]{#1}
\csname url@samestyle\endcsname
\providecommand{\newblock}{\relax}
\providecommand{\bibinfo}[2]{#2}
\providecommand{\BIBentrySTDinterwordspacing}{\spaceskip=0pt\relax}
\providecommand{\BIBentryALTinterwordstretchfactor}{4}
\providecommand{\BIBentryALTinterwordspacing}{\spaceskip=\fontdimen2\font plus
\BIBentryALTinterwordstretchfactor\fontdimen3\font minus
  \fontdimen4\font\relax}
\providecommand{\BIBforeignlanguage}[2]{{%
\expandafter\ifx\csname l@#1\endcsname\relax
\typeout{** WARNING: IEEEtran.bst: No hyphenation pattern has been}%
\typeout{** loaded for the language `#1'. Using the pattern for}%
\typeout{** the default language instead.}%
\else
\language=\csname l@#1\endcsname
\fi
#2}}
\providecommand{\BIBdecl}{\relax}
\BIBdecl

\bibitem{panno2021review}
\BIBentryALTinterwordspacing
S.~Panno, S.~Davino, A.~G. Caruso, S.~Bertacca, A.~Crnogorac, A.~Mandi{\'c},
  E.~Noris, and S.~Mati{\'c}, ``A review of the most common and economically
  important diseases that undermine the cultivation of tomato crop in the
  mediterranean basin,'' \emph{Agronomy}, vol.~11, no.~11, p. 2188, 2021.
  [Online]. Available: \url{https://www.mdpi.com/2073-4395/11/11/2188}
\BIBentrySTDinterwordspacing

\bibitem{agriculture11080707}
\BIBentryALTinterwordspacing
J.~Lu, L.~Tan, and H.~Jiang, ``Review on convolutional neural network (cnn)
  applied to plant leaf disease classification,'' \emph{Agriculture}, vol.~11,
  no.~8, 2021. [Online]. Available:
  \url{https://www.mdpi.com/2077-0472/11/8/707}
\BIBentrySTDinterwordspacing

\bibitem{dhaka2021aSurvey}
\BIBentryALTinterwordspacing
V.~S. Dhaka, S.~V. Meena, G.~Rani, D.~Sinwar, Kavita, M.~F. Ijaz, and
  M.~Woźniak, ``A survey of deep convolutional neural networks applied for
  prediction of plant leaf diseases,'' \emph{Sensors}, vol.~21, no.~14, 2021.
  [Online]. Available: \url{https://www.mdpi.com/1424-8220/21/14/4749}
\BIBentrySTDinterwordspacing

\bibitem{s18082674}
\BIBentryALTinterwordspacing
K.~G. Liakos, P.~Busato, D.~Moshou, S.~Pearson, and D.~Bochtis, ``Machine
  learning in agriculture: A review,'' \emph{Sensors}, vol.~18, no.~8, 2018.
  [Online]. Available: \url{https://www.mdpi.com/1424-8220/18/8/2674}
\BIBentrySTDinterwordspacing

\bibitem{kamilaris2018deep}
\BIBentryALTinterwordspacing
A.~Kamilaris and F.~X. Prenafeta-Boldú, ``Deep learning in agriculture: A
  survey,'' \emph{Computers and Electronics in Agriculture}, vol. 147, pp.
  70--90, 2018. [Online]. Available:
  \url{https://www.sciencedirect.com/science/article/pii/S0168169917308803}
\BIBentrySTDinterwordspacing

\bibitem{kamal2022huruf}
\BIBentryALTinterwordspacing
M.~Kamal, F.~Shaiara, C.~M. Abdullah, S.~Ahmed, T.~Ahmed, and M.~H. Kabir,
  ``Huruf: An application for arabic handwritten character recognition using
  deep learning,'' 2022. [Online]. Available:
  \url{https://arxiv.org/abs/2212.08610}
\BIBentrySTDinterwordspacing

\bibitem{alamgir2022dhakaAI}
\BIBentryALTinterwordspacing
R.~M. Alamgir, A.~A. Shuvro, M.~A. Mushabbir, M.~A. Raiyan, N.~J. Rani, M.~M.
  Rahman, M.~H. Kabir, and S.~Ahmed, ``Performance analysis of yolo-based
  architectures for vehicle detection from traffic images in bangladesh,''
  2022. [Online]. Available: \url{https://arxiv.org/abs/2212.09144}
\BIBentrySTDinterwordspacing

\bibitem{sakib2021bdslReview}
\BIBentryALTinterwordspacing
A.~Khatun, M.~S. Shahriar, M.~H. Hasan, K.~Das, S.~Ahmed, and M.~S. Islam, ``A
  systematic review on the chronological development of bangla sign language
  recognition systems,'' in \emph{2021 Joint 10th International Conference on
  Informatics, Electronics \& Vision (ICIEV) and 2021 5th International
  Conference on Imaging, Vision \& Pattern Recognition (icIVPR)}, 2021, pp.
  1--9. [Online]. Available:
  \url{https://ieeexplore.ieee.org/abstract/document/9564157}
\BIBentrySTDinterwordspacing

\bibitem{bakhtiar2022traffic}
\BIBentryALTinterwordspacing
R.~Rahman, Z.~Bin~Azad, and M.~Bakhtiar~Hasan, ``Densely-populated traffic
  detection using yolov5 and non-maximum suppression ensembling,'' in
  \emph{Proceedings of the International Conference on Big Data, IoT, and
  Machine Learning}.\hskip 1em plus 0.5em minus 0.4em\relax Springer Singapore,
  2022, pp. 567--578. [Online]. Available:
  \url{https://link.springer.com/chapter/10.1007/978-981-16-6636-0_43}
\BIBentrySTDinterwordspacing

\bibitem{bakhtiar2022heatgait}
\BIBentryALTinterwordspacing
M.~B. Hasan, T.~Ahmed, and M.~H. Kabir, ``Heatgait: Hop-extracted adjacency
  technique in graph convolution based gait recognition,'' in \emph{2022 4th
  International Conference on Advances in Computer Technology, Information
  Science and Communications (CTISC)}, 2022, pp. 1--6. [Online]. Available:
  \url{https://ieeexplore.ieee.org/abstract/document/9849799}
\BIBentrySTDinterwordspacing

\bibitem{khan2022rethinking}
\BIBentryALTinterwordspacing
A.~M. Khan, A.~Ashrafee, R.~Sayera, S.~Ivan, and S.~Ahmed, ``Rethinking cooking
  state recognition with vision transformers,'' 2022. [Online]. Available:
  \url{https://arxiv.org/abs/2212.08586}
\BIBentrySTDinterwordspacing

\bibitem{mohanty2016using}
\BIBentryALTinterwordspacing
S.~P. Mohanty, D.~P. Hughes, and M.~Salath{\'e}, ``Using deep learning for
  image-based plant disease detection,'' \emph{Frontiers in plant science},
  vol.~7, p. 1419, 2016. [Online]. Available:
  \url{https://www.frontiersin.org/articles/10.3389/fpls.2016.01419}
\BIBentrySTDinterwordspacing

\bibitem{sethy2020deep}
\BIBentryALTinterwordspacing
P.~K. Sethy, N.~K. Barpanda, A.~K. Rath, and S.~K. Behera, ``Deep feature based
  rice leaf disease identification using support vector machine,''
  \emph{Computers and Electronics in Agriculture}, vol. 175, p. 105527, 2020.
  [Online]. Available:
  \url{https://www.sciencedirect.com/science/article/abs/pii/S0168169919326997}
\BIBentrySTDinterwordspacing

\bibitem{mishra2020deep}
\BIBentryALTinterwordspacing
S.~Mishra, R.~Sachan, and D.~Rajpal, ``Deep convolutional neural network based
  detection system for real-time corn plant disease recognition,''
  \emph{Procedia Computer Science}, vol. 167, pp. 2003--2010, 2020. [Online].
  Available:
  \url{https://www.sciencedirect.com/science/article/pii/S187705092030702X}
\BIBentrySTDinterwordspacing

\bibitem{ahmed2022less}
\BIBentryALTinterwordspacing
S.~Ahmed, M.~B. Hasan, T.~Ahmed, M.~R.~K. Sony, and M.~H. Kabir, ``Less is
  more: lighter and faster deep neural architecture for tomato leaf disease
  classification,'' \emph{IEEE Access}, vol.~10, pp. 68\,868--68\,884, 2022.
  [Online]. Available:
  \url{https://ieeexplore.ieee.org/abstract/document/9810234}
\BIBentrySTDinterwordspacing

\bibitem{tiwari2020potato}
\BIBentryALTinterwordspacing
D.~Tiwari, M.~Ashish, N.~Gangwar, A.~Sharma, S.~Patel, and S.~Bhardwaj,
  ``Potato leaf diseases detection using deep learning,'' in \emph{2020 4th
  International Conference on Intelligent Computing and Control Systems
  (ICICCS)}.\hskip 1em plus 0.5em minus 0.4em\relax IEEE, 2020, pp. 461--466.
  [Online]. Available:
  \url{https://ieeexplore.ieee.org/abstract/document/9121067}
\BIBentrySTDinterwordspacing

\bibitem{liu2017identification}
\BIBentryALTinterwordspacing
B.~Liu, Y.~Zhang, D.~He, and Y.~Li, ``Identification of apple leaf diseases
  based on deep convolutional neural networks,'' \emph{Symmetry}, vol.~10,
  no.~1, p.~11, 2017. [Online]. Available:
  \url{https://www.mdpi.com/2073-8994/10/1/11}
\BIBentrySTDinterwordspacing

\bibitem{ramedani2014potential}
\BIBentryALTinterwordspacing
Z.~Ramedani, M.~Omid, A.~Keyhani, S.~Shamshirband, and B.~Khoshnevisan,
  ``Potential of radial basis function based support vector regression for
  global solar radiation prediction,'' \emph{Renewable and Sustainable Energy
  Reviews}, vol.~39, pp. 1005--1011, 2014. [Online]. Available:
  \url{https://www.sciencedirect.com/science/article/abs/pii/S1364032114005607}
\BIBentrySTDinterwordspacing

\bibitem{chuanlei2017apple}
\BIBentryALTinterwordspacing
Z.~Chuanlei, Z.~Shanwen, Y.~Jucheng, S.~Yancui, and C.~Jia, ``Apple leaf
  disease identification using genetic algorithm and correlation based feature
  selection method,'' \emph{International Journal of Agricultural and
  Biological Engineering}, vol.~10, no.~2, pp. 74--83, 2017. [Online].
  Available: \url{http://www.ijabe.org/index.php/ijabe/article/view/2166}
\BIBentrySTDinterwordspacing

\bibitem{jiang2019real}
\BIBentryALTinterwordspacing
P.~Jiang, Y.~Chen, B.~Liu, D.~He, and C.~Liang, ``Real-time detection of apple
  leaf diseases using deep learning approach based on improved convolutional
  neural networks,'' \emph{IEEE Access}, vol.~7, pp. 59\,069--59\,080, 2019.
  [Online]. Available:
  \url{https://ieeexplore.ieee.org/abstract/document/8706936}
\BIBentrySTDinterwordspacing

\bibitem{zhong2020research}
\BIBentryALTinterwordspacing
Y.~Zhong and M.~Zhao, ``Research on deep learning in apple leaf disease
  recognition,'' \emph{Computers and Electronics in Agriculture}, vol. 168, p.
  105146, 2020. [Online]. Available:
  \url{https://www.sciencedirect.com/science/article/abs/pii/S016816991931556X}
\BIBentrySTDinterwordspacing

\bibitem{baranwal2019deep}
\BIBentryALTinterwordspacing
S.~Baranwal, S.~Khandelwal, and A.~Arora, ``Deep learning convolutional neural
  network for apple leaves disease detection,'' in \emph{Proceedings of
  International Conference on Sustainable Computing in Science, Technology and
  Management (SUSCOM), Amity University Rajasthan, Jaipur-India}, 2019.
  [Online]. Available:
  \url{https://papers.ssrn.com/sol3/papers.cfm?abstract_id=3351641}
\BIBentrySTDinterwordspacing

\bibitem{al2020evolutionary}
\BIBentryALTinterwordspacing
J.~S.~H. Al-bayati and B.~B. {\"U}st{\"u}nda{\u{g}}, ``Evolutionary feature
  optimization for plant leaf disease detection by deep neural networks,''
  \emph{International Journal of Computational Intelligence Systems}, vol.~13,
  no.~1, p.~12, 2020. [Online]. Available:
  \url{https://www.atlantis-press.com/journals/ijcis/125931940}
\BIBentrySTDinterwordspacing

\bibitem{li2020apple}
\BIBentryALTinterwordspacing
X.~Li and L.~Rai, ``Apple leaf disease identification and classification using
  resnet models,'' in \emph{2020 IEEE 3rd International Conference on
  Electronic Information and Communication Technology (ICEICT)}.\hskip 1em plus
  0.5em minus 0.4em\relax IEEE, 2020, pp. 738--742. [Online]. Available:
  \url{https://ieeexplore.ieee.org/abstract/document/9334214}
\BIBentrySTDinterwordspacing

\bibitem{agarwal2019fcnn}
\BIBentryALTinterwordspacing
M.~Agarwal, R.~K. Kaliyar, G.~Singal, and S.~K. Gupta, ``Fcnn-lda: a faster
  convolution neural network model for leaf disease identification on apple's
  leaf dataset,'' in \emph{2019 12th International Conference on Information \&
  Communication Technology and System (ICTS)}.\hskip 1em plus 0.5em minus
  0.4em\relax IEEE, 2019, pp. 246--251. [Online]. Available:
  \url{https://ieeexplore.ieee.org/document/8850964}
\BIBentrySTDinterwordspacing

\bibitem{tan2021efficientnetv2}
\BIBentryALTinterwordspacing
M.~Tan and Q.~Le, ``Efficientnetv2: Smaller models and faster training,'' in
  \emph{International Conference on Machine Learning}.\hskip 1em plus 0.5em
  minus 0.4em\relax PMLR, 2021, pp. 10\,096--10\,106. [Online]. Available:
  \url{http://proceedings.mlr.press/v139/tan21a.html}
\BIBentrySTDinterwordspacing

\bibitem{hughes2015open}
\BIBentryALTinterwordspacing
D.~P. Hughes and M.~Salath{\'{e}}, ``{An open access repository of images on
  plant health to enable the development of mobile disease diagnostics through
  machine learning and crowdsourcing},'' \emph{{arXiv - Computing Research
  Repository}}, 2015. [Online]. Available:
  \url{http://arxiv.org/abs/1511.08060}
\BIBentrySTDinterwordspacing

\bibitem{tasnim2019bangla}
\BIBentryALTinterwordspacing
T.~Ahmed, M.~N. Raihan, R.~Kushol, and M.~S. Salekin, ``A complete bangla
  optical character recognition system: An effective approach,'' in \emph{2019
  22nd International Conference on Computer and Information Technology
  (ICCIT)}, 2019, pp. 1--7. [Online]. Available:
  \url{https://ieeexplore.ieee.org/abstract/document/9038551/}
\BIBentrySTDinterwordspacing

\bibitem{morshed2022fruit}
\BIBentryALTinterwordspacing
M.~S. Morshed, S.~Ahmed, T.~Ahmed, M.~U. Islam, and A.~B. M.~A. Rahman, ``Fruit
  quality assessment with densely connected convolutional neural network,''
  2022. [Online]. Available: \url{https://arxiv.org/abs/2212.04255}
\BIBentrySTDinterwordspacing

\bibitem{yasmeen2021csvcNet}
\BIBentryALTinterwordspacing
A.~Yasmeen, F.~I. Rahman, S.~Ahmed, and M.~H. Kabir, ``Csvc-net: Code-switched
  voice command classification using deep cnn-lstm network,'' in \emph{2021
  Joint 10th International Conference on Informatics, Electronics \& Vision
  (ICIEV) and 2021 5th International Conference on Imaging, Vision \& Pattern
  Recognition (icIVPR)}, 2021, pp. 1--8. [Online]. Available:
  \url{https://ieeexplore.ieee.org/abstract/document/9564183}
\BIBentrySTDinterwordspacing

\bibitem{ashikur2022twoDecades}
\BIBentryALTinterwordspacing
A.~B.~M. Ashikur~Rahman, M.~B. Hasan, S.~Ahmed, T.~Ahmed, M.~H. Ashmafee, M.~R.
  Kabir, and M.~H. Kabir, ``Two decades of bengali handwritten digit
  recognition: A survey,'' \emph{IEEE Access}, vol.~10, pp. 92\,597--92\,632,
  2022. [Online]. Available:
  \url{https://ieeexplore.ieee.org/abstract/document/9869842}
\BIBentrySTDinterwordspacing

\bibitem{Sandler_2018_CVPR}
\BIBentryALTinterwordspacing
M.~Sandler, A.~Howard, M.~Zhu, A.~Zhmoginov, and L.-C. Chen, ``Mobilenetv2:
  Inverted residuals and linear bottlenecks,'' in \emph{Proceedings of the IEEE
  Conference on Computer Vision and Pattern Recognition (CVPR)}, June 2018.
  [Online]. Available:
  \url{https://openaccess.thecvf.com/content_cvpr_2018/html/Sandler_MobileNetV2_Inverted_Residuals_CVPR_2018_paper.html}
\BIBentrySTDinterwordspacing

\bibitem{batchnormalization}
\BIBentryALTinterwordspacing
S.~Ioffe and C.~Szegedy, ``Batch normalization: Accelerating deep network
  training by reducing internal covariate shift,'' in \emph{Proceedings of the
  32nd International Conference on Machine Learning}, ser. Proceedings of
  Machine Learning Research, F.~Bach and D.~Blei, Eds., vol.~37.\hskip 1em plus
  0.5em minus 0.4em\relax Lille, France: PMLR, 07--09 Jul 2015, pp. 448--456.
  [Online]. Available: \url{https://proceedings.mlr.press/v37/ioffe15.html}
\BIBentrySTDinterwordspacing

\bibitem{dropout}
\BIBentryALTinterwordspacing
N.~Srivastava, G.~Hinton, A.~Krizhevsky, I.~Sutskever, and R.~Salakhutdinov,
  ``Dropout: A simple way to prevent neural networks from overfitting,''
  \emph{Journal of Machine Learning Research}, vol.~15, no.~56, pp. 1929--1958,
  2014. [Online]. Available:
  \url{http://jmlr.org/papers/v15/srivastava14a.html}
\BIBentrySTDinterwordspacing

\end{thebibliography}

\end{document}